\documentclass[10pt, conference, compsocconf]{IEEEtran}
\usepackage[pdftex]{graphicx}
\usepackage{amsmath, amssymb, subfig, caption}
\usepackage{eqparbox}
\usepackage{multirow,tabulary,booktabs}
\usepackage{makecell, booktabs, multirow, siunitx, xcolor}

\newcommand{\RN}[1]{%
  \textup{\uppercase\expandafter{\romannumeral#1}}%
}

\begin{document}
%
\title{A Concise Review of Transfer Learning}

\makeatletter
\newcommand{\linebreakand}{
  \end{@IEEEauthorhalign}
  \hfill\mbox{}\par
  \mbox{}\hfill\begin{@IEEEauthorhalign}
}
\makeatother
\author{\IEEEauthorblockN{Abolfazl Farahani}
\IEEEauthorblockA{Department of Computer Science\\
University of Georgia\\
Athens, GA, USA \\
a.farahani@uga.edu}
\and
\IEEEauthorblockN{Behrouz Pourshojae}
\IEEEauthorblockA{Department of Information and Technology\\
Road and Urban Development Organization\\
Arak, Iran \\
b.pourshojae@gmail.com}
\and
\IEEEauthorblockN{Khaled Rasheed}
\IEEEauthorblockA{Department of Computer Science\\
University of Georgia\\
Athens, GA, USA \\
Khaled@uga.edu}
\linebreakand 
\IEEEauthorblockN{Hamid R. Arabnia}
\IEEEauthorblockA{Department of Computer Science\\
University of Georgia\\
Athens, GA, USA \\
hra@uga.edu}
}


%


\maketitle
\thispagestyle{plain}
\pagestyle{plain}

\begin{abstract}
     The availability of abundant labeled data in recent years led the researchers to introduce a methodology called transfer learning, which utilizes existing data in situations where there are difficulties in collecting new annotated data. Transfer learning aims to boost the performance of a target learner by applying another related source data.
     In contrast to the traditional machine learning and data mining techniques, which assume that the training and testing data lie from the same feature space and distribution, transfer learning can handle situations where there is a discrepancy between domains and distributions. These characteristics give the model the potential to utilize the available related source data and extend the underlying knowledge to the target task achieving better performance.
    This survey paper aims to give a concise review of traditional and current transfer learning settings, existing challenges, and related approaches.

\end{abstract}


%
\IEEEpeerreviewmaketitle

\section{Introduction}
Machine learning technologies have already been applied in a wide verity of real-world applications and enjoyed significant success. 
Traditional machine learning models are trained on a series of labeled samples from the existing collected data called training data, and later these trained models can be applied to predict the label for the new unseen data termed as test data. 
The prosperity of traditional machine learning techniques highly depends on many training examples and is restricted by the assumption of having the same domain space and distribution in both training and testing data. However, in real-world applications, there are situations in which the training and testing instances originate from different feature spaces or distributions. It usually happens due to the difficulty of collecting new test instances with the same property, dimension, and distribution as we have in the existing training data. Thus, if the above assumption is violated, the model trained on the training data will fail to perform well on the test data, and the model needs to be trained from scratch for the new data. Training the model from scratch or collecting new annotated training data is expensive, time-consuming, and sometimes impossible. Besides, it is more convenient to utilize the available large volume of labeled data for training a learner that is capable of being trained on one domain with easily collected data and applied to another related domain with scarce data. This methodology is called transfer learning, which is motivated by the fact that humans can learn faster, easier, and more efficiently using the knowledge from previously learned tasks. In this methodology, domains, tasks, and distributions can vary between training and testing data while they are related in some ways. 
Transfer learning and domain adaptation has attracted much attention and actively researched in the past decade. Domain adaptation \cite{csurka2017comprehensive}, \cite{farahani2020brief} is a subset of transfer learning where the tasks are equivalent between domains.
Transfer learning aims to improve a target model's performance with insufficient or lack of annotated data by using the knowledge from another related source domain with adequate labeled data. This model is assumed to predict the label for a new unseen target data more accurately as compared to a model trained only with limited available target data.

This paper intends to briefly review transfer learning, its categories, and some current solutions to each category. Therefore, it is not meant to be a comprehensive survey paper; instead, it is a concise review in the sense that only a few representative transfer learning approaches from each intertwined/similar group of settings are considered.

\section{Related Work}
In this section, we address some machine learning techniques that are closely related to transfer learning.
    
\textit{Semi-supervised learning} \cite{zhu2005semi}, \cite{pise2008survey} is a machine learning technique that addresses the problems where labeled data is not available adequately. It employs abundant unlabeled samples and a small amount of annotated samples to train a model. Semi-supervised learning assumes that both labeled and unlabeled data in the training and test sets are distributed independently and identically. However, this assumption does not hold in transfer learning; hence, the training and testing data are allowed to be drawn from distinct domains, tasks, and distributions.
    
\textit{Multi-view learning} \cite{sun2013survey} is another related task aiming to learn from multi-view data or multiple sets of distinctive features such as audio and video signals, image and text data, or even two different text documents that provide different explanations about a subject. For example, a web page can be represented by a text document and an anchor text in its hyperlink. 
Multi-view data contains complementary information that helps the model to learn more informative and comprehensive representations.
Multi-view learning has been used broadly in many real-world applications due to the availability of multi-view data in recent years. 
Natural language processing, video analysis, recommender system, and cross-media retrieval are some applications that benefit from this type of learning. 
Canonical correlation analysis (CCA) \cite{hotelling1992relations} and co-training \cite{blum1998combining} are the first representative techniques introduced in the concept of multi-view learning which also used in transfer learning \cite{bo2015transfer}.

Similar to transfer learning, \textit{multi-task learning} \cite{zhang2017survey} tends to enhance the performance of a learner using knowledge transfer. It simultaneously trains multiple related tasks to improve their generalization. 
The assumption is that related tasks share common information and utilize the same knowledge. Therefore, multi-task learning aims to learn the underlying structure of data and share it between all the tasks. Transfer learning and multi-task learning both employ parameter sharing and feature transformation for the knowledge transfer. The difference between these two learning strategies is that the multi-task learning aims to improve the performance in all related tasks, while transfer learning focuses only on the target learner and tends to boost its performance.  

\section{Overview}
In the following section, the notations and definitions used in this survey are addressed. Besides, the categorizations of transfer learning are clearly introduced from different viewpoints.
    \subsection{Notations and Definitions}
        \textbf{\textit{Definition 1 (Domain).}} A domain \(\mathcal{D}=\{\mathcal{X}, P(X)\}\) consists of two components; a feature space \(\mathcal{X}\) , and a marginal probability distribution \(P(X)\), where \(X\) is defined as an instance set,  and \(X = \{ x_1, x_2,..., x_n\}\in \mathcal{X}\).
        
        \textbf{\textit{Definition 2 (Task).}}
            A task \(\mathcal{T}\) composed of a label space \(\mathcal{Y}\) and an objective predictive function \(f(.)\). I.e., \(\mathcal{T}=\{\mathcal{Y},f(.)\}\).  Given a specific domain \(\mathcal{D}=\{\mathcal{X}, P(X)\}\), the sample data consists of pairs \(\{x_i,y_i\}\) where \(x_i\in X\) and \(y_i\in \mathcal{Y}\). The objective function \(f\) is supposed to learn from sample data to predict the corresponding label for the new instances. In other word, \(f\) is considered as conditional distribution of instances and can be written as \(f(x)=P(y|x)\).
            
        \textbf{\textit{Definition 3 (Transfer Learning).}}
            Given a source domain \(\mathcal{D}_S\) with corresponding source task \(\mathcal{T}_S\), and target domain \(\mathcal{D}_T\) with corresponding target task \(\mathcal{T}_T\), transfer learning aims to transfer the related knowledge contained in \(\mathcal{D}_S\) and \(\mathcal{T}_S\) to boost the performance of the target predictive function \(f_T(.)\) in the target task \(\mathcal{T}_T\) and target domain \(\mathcal{D}_T\) where \(\mathcal{D}_S \neq \mathcal{D}_T\) or \(\mathcal{T}_S \neq \mathcal{T}_T\).
            
        The above definitions belong to the single-source transfer learning, which is more considered in the existing transfer learning studies. However, there are some studies regarding multi-source transfer learning \cite{fang2015multi} in which multiple source domains and tasks contribute to improving the predictive function  \( f_T \).
        Transfer learning relaxes the assumption of having equivalent source and target domains and tasks that holds in the classical machine learning problems. The changes in domains and tasks make different scenarios that can be elaborated as follows.
        Since domain is defined as a pair \(\mathcal{D}=\{\mathcal{X}, P(X)\}\), different domains, \(\mathcal{D}_S \neq \mathcal{D}_T\), indicates the situations where either \(\mathcal{X}_S \neq \mathcal{X}_T\) or \(P(X_S) \neq P(X_T)\) or both might be the case. When domains are drawn from different feature spaces, \(\mathcal{X}_S \neq \mathcal{X}_T\), they are still supposed to share common semantics in the latent space. 
        When marginal distributions are different between domains, i.e., \(P(X_S) \neq P(X_T)\), transfer learning tends to reduce the discrepancy between domains. E.g., having city street images taken in the daylight and at night as a source domain and target domain, respectively. Another example is to have documents with different topics as the source and target domains. This setting is also referred to as frequency feature bias.
        When both feature space and marginal distribution vary between domains, it implies that the samples in both domains are related, but they are represented in different ways, e.g., having two documents with different languages and topics as the source domain and target domain. In this case, knowledge transfer is more challenging.
        Regarding the definition of task, \(\mathcal{T}=\{\mathcal{Y},P(Y|X)\}\), different tasks, \(\mathcal{T}_S \neq \mathcal{T}_T\), indicates that either label spaces are different, \(Y_S \neq Y_T\), or conditional probability distributions vary between the source and target, \(P(Y_S|X_S) \neq P(Y_T|X_T)\). Binary class classification as a source task and multi-class classification as a target task is an example of having different label spaces. 
        An example of different conditional distributions is when a word has different meanings in various documents, e.g., word \textit{small} has a positive meaning when referring to the size of an electronic gadget and a negative meaning when talking bout space in a car. Different conditional distributions are also termed as context feature bias.
        \(P(Y_s) \neq P(Y_T)\) is another possible setting when the labeled data in the source and target domains are unbalanced.
        
        All of the above transfer learning settings need to be treated differently. In the following sections, we will discuss some of the approaches introduced to deal with each scenario.
    \subsection{Categorization of Transfer Learning}
        The categorization of transfer learning varies based on different viewpoints and characteristics. Traditional categorization of transfer learning is presented by \cite{pan2010survey}.
        Based on the availability and the size of labeled data in the source and target domains, \cite{pan2010survey} divides the transfer learning problems into three main categories, i.e., inductive, transductive, and unsupervised transfer learning.
        
        Inductive transfer learning indicates the situation that labeled data is available in the target domain. In this setting, the source and target tasks are considered to be different while the source and target domains might be similar or different.
        In transductive transfer learning, source and target domains are different; however, the source and target tasks are the same. In this setting, the source domain contains a large amount of labeled data while there is no labeled data in the target domain.
        Unsupervised Transfer learning refers to the situation in which neither source domain nor target domain contains labeled data.
        
        We also can categorize transfer learning into homogeneous and heterogeneous settings regarding different scenarios of feature space and label space across domains. Homogeneous transfer learning problems refer to the situation where the source and target domains share identical feature and label spaces while domain divergence exists due to the marginal or/and conditional distribution differences between domains.
        When the feature or/and label spaces varies between domains, the problem is considered as Heterogeneous transfer leaning.
        
        In the next section, we will address some of the approaches that have been proposed to solve the homogeneous and heterogeneous transfer learning problems.
\section{Approach}
    Solutions to the homogeneous and heterogeneous transfer learning problem settings can be generally divided into shallow and deep transfer learning approaches. Classical approaches are considered shallow transfer learning and utilize instance-based, feature-based, parameter-based, and relational-based \cite{pan2010survey} methods. In addition to the above methods, \cite{weiss2016survey} considers the hybrid-based approach that benefits the combination of instance-based and parameter-based algorithms. On the other hand, deep networks have recently gained much attention in the machine learning community and have been applied in many applications such as Natural Language Processing \cite{zhang2018deep}, image analysis \cite{amirian2018}, \cite{mohammadi2017region}, video analysis \cite{toutiaee2020video}, speech recognition \cite{dahl2011context}, \cite{asali2020deepmsrf}, and many others. \nocite{*} The growth of available annotated data and the success of deep learning in discovering the underlying structure of mass data in the past decade led the researcher to introduce a new methodology known as deep transfer learning. This approach aims to utilize deep learning techniques to address transfer learning problems.
    
    In this paper, we only discuss some of the shallow transfer learning approaches.
\subsection{Homogeneous Transfer Learning}
    Homogeneous transfer learning techniques are used to handle situations where the feature and label spaces are the same across domains, while the marginal or/and conditional distributions may differ. The homogeneous transfer learning approaches aim to diminish the distribution difference between domains. They mainly solve the problem by matching the marginal distributions, the conditional distributions, or both. They usually utilize different metrics such as Maximum Mean Discrepancy (MMD) \cite{gretton2007kernel}, Kullback-Leibler (KL) divergence \cite{kullback1951information}, and correlation alignment (CORAL) \cite{sun2017correlation} to measure the domain divergence and then minimize this discrepancy. Instance-based, feature-based, parameter-based, and relational-based are the main categories of homogeneous transfer learning approaches. In this section, we briefly discuss each category and address some of the existing techniques accordingly. 
    \subsubsection{Instance-Based}
        Instance-based approaches tend to minimize the marginal or/and conditional distribution difference between domains using importance sampling or re-weighting methods. This strategy mainly aims to find proper weights for the source labeled data to learn the source task with minimum expected risk when applying to the target domain. For further information about risk minimization, see \cite{vapnik1992principles}. Therefore, The expected risk of the target task can be written based on the source distribution.
        \begin{gather*}
            \mathbb{E}_{(x,y)\sim P_T}\left[\ell(f(x),y)\right]= \mathbb{E}_{(x,y)\sim P_S}\left[\frac{p_\mathcal{T}(x,y)}{p_\mathcal{S}(x,y)}\ell(f(x),y)\right],
        \end{gather*}
        where $P_S(x,y)$ and $P_T(x,y)$ are the source and target joint probability distributions, respectively and the ratio of these density functions, $\frac{p_\mathcal{T}(x,y)}{p_\mathcal{S}(x,y)}$, is the weighting parameter.
        $\ell(f(x),y)$ is a loss function indicating the disagreement between the predicted label by predictive function $f$ and the ground truth. Regarding the labeled samples in the source domain the above expected risk can be approximated by the sample average in the source domain,
        \begin{equation*}
        \small
        \begin{split}
             \mathbb{E}_{(x,y)\sim P_T}\left[\ell(f(x),y)\right]&=\underset{f}{\mathrm{argmin}}\:\mathbb{E}_{(x,y)\sim P_S}\left[\frac{p_\mathcal{T}(x,y)}{p_\mathcal{S}(x,y)}\ell(f(x),y)\right]\\                    &\approx\underset{f}{\mathrm{argmin}}\:\frac{1}{n_S}\sum\limits_{i=1}^{n_S}\beta_i\ell(f(x_i),y_i)+\Omega(f),
        \end{split}
        \end{equation*}
        where $\beta=\frac{p_\mathcal{T}(x,y)}{p_\mathcal{S}(x,y)}$ is the weighting parameter for each source instance, and $\Omega$ is the regularization term. Thus, proper weights are computed as the penalty for the source samples to learn a model with minimum risk on the target domain. $\beta$ can be achieved by indirectly estimating the source and target probability distributions individually and then computing the ratio. However, this approach is very challenging and ineffective \cite{cortes2010learning}. Therefore, the weights can be estimated directly in an optimization procedure, where the model minimizes the discrepancy between the distributions by re-weighting the samples. Kernel Mean Matching (KMM) \cite{huang2007correcting} utilizes Maximum Mean Discrepancy (MMD) to minimizes the marginal distribution difference between domains. KMM obtains the source sample weights by minimizing the empirical means of source and target distributions in a Reproducing Kernel Hilbert Space (RKHS),
        \begin{gather*}
        \underset{\beta_i \in [0,B]}{\mathrm{argmin}}\quad||\frac{1}{m}\sum\limits_{i=1}^{m}\beta_i\Phi(x_i)-\frac{1}{m'}\sum\limits_{i=1}^{m'}\Phi(x'_i)||^2\\
        \textrm{s.t.} \quad |\frac{1}{m}\sum\limits_{i=1}^{m}\beta_i-1|\leq \epsilon \quad \textrm{and}\quad 0\leq \beta_i \leq B,
        \end{gather*}
        where $\beta$ is the weighting parameter, and $\Phi$ represents the mapping function. $m$ and $m'$ are the number of data samples in the source and target domains respectively. In the second line of the equation, $\epsilon$ is a small parameter, and $B$ is a parameter for constraint.
        \cite{chattopadhyay2012multisource} introduces two different techniques called Conditional Probability-based Multi-source Domain Adaptation (CP-MDA) and two-stage weighting framework for multi-source domain adaptation (2SW-MDA) to minimize the discrepancy between distributions.
        The proposed frameworks consider multi-source labeled data with different distributions along with limited labeled and a large amount of unlabeled data in the target domain.
        CP-MDA is based on the hypothesis weighting that aims to match the conditional distribution differences between domains. To do so, it trains individual classifiers for each source domain, and a weight value is then computed for each source classifier based on the similarity between the specific source and target domain distributions. In the next step, all the weighted source classifiers are combined to predict the pseudo labels for unlabeled data in the target domain. Finally, the target classifier can learn from both the limited labeled and pseudo labeled data in the target domain.
        2SW-MDA reduces the marginal and conditional distributions between multi-source and target domains. In the first stage, It finds sample weights for the source domain data to match the marginal distributions between each source domain and the target domain. 2SW-MDA utilizes Maximum Mean Discrepancy (MMD) to minimize the discrepancy between means of source and target distributions in a Reproduced Kernel Hilbert Space (RKHS). Similar to CP-MDA, the second stage computes the weights for each source domain based on their similarity to the target domain to match the conditional distribution discrepancy. 
        Eventually, the target classifier can be trained on the re-weighted source samples and a few labeled target samples (if available). 
    \subsubsection{Feature-Based}
        Homogeneous feature-based transfer learning approaches attempt to minimize the marginal or/and conditional distribution discrepancy between domains. They usually transforms the original instances into a new space to discover the related underlying structures across domains. Approaches in this category apply various techniques such as feature augmentation, feature mapping, feature clustering, feature Alignment, feature endcoding, and feature selection.
        Feature-based transfer learning can be generally divided into two categories; symmetric feature transformation and asymmetric feature transformation.
        \paragraph{Symmetric Feature-Based}
            This transfer learning strategy aims to discover good and common features across domains in the latent space. Common features are assumed to help the learner to boost the performance in the target domain.
            Transfer component analysis (TCA) \cite{pan2010domain} is a feature mapping technique proposed to deal with the situations where the labeled data is not available in the target domain. This approach aims to discover domain-invariant feature representation across domains by minimizing the difference between marginal distributions. TCA uses Maximum Mean Discrepancy (MMD) to measure the discrepancy between marginal distributions in a reproducing kernel Hilbert space (RKHS). These common features can be used by any machine learning method to learn the final target classifier.
            Sampling Geodesic Flow (SGF) \cite{gopalan2011domain} is a feature alignment method that tends to discover a common feature representation across domains in a  shared low-dimensional Grassmann manifold. In this technique, the source and target domains are constructed as two subspaces in the Grassmann manifold space using a dimensionality reduction technique. Next, SGF finds a geodesic path between the source and target points and samples a set of subspaces from this path.  Data from both domains are then projected onto all the sampled subspaces and are concatenated to construct two high-dimensional vectors, one for each domain. Finally, We can train a classifier on the high-dimensional source vector and apply it to the target vector to predict the target labels. We can boost the performance by sampling more points from the geodesic path. However, it increases the dimensionality of the feature vectors, which is computationally expensive and impractical. 
            Geodesic Flow Kernel \cite{gong2012geodesic}, was proposed to extend and improve SGF. GFK is a kernel-based domain adaptation method that uses the feature alignment strategy to deals with the domain shift. It aims to represent the smoothness of transition from a source to a target domain by integrating an infinite number of subspaces to find a geodesic line between domains in a low-dimensional manifold space.
        \paragraph{Asymmetric Feature-Based}
            Asymmetric feature-based approaches aim to reduce the discrepancy between domain distributions by transforming the source domain features into the target domain. The transformation computes new weights for the source samples such that both distributions are aligned in the target space.
            When the conditional distribution difference is the reason for domain discrepancy, the trained source classifier cannot be applied directly to the target domain since it will harm the performance. Conditional distribution discrepancy happens when the features have different meanings across domains. This issue is known as context feature bias. E.g., In the text classification problem, context feature bias refers to the situation where the source and target domains contain different topics meaning that some words in the source domain have different meanings than those in the target domain. 
            Feature Augmentation Method (FAM) \cite{daume2009frustratingly} was proposed to address the above issue in Natural Language Processing (NLP) problems. This technique assumes that the source and target domains are similar, i.e.,${\mathcal{D}_s\sim \mathcal{D}_t}$ while the conditional distributions vary between domains. FAM utilizes a simple feature augmentation strategy to create two new sets of instances for the source and target domains. The new feature space's size is three times its original size, consisting of three copies of the original features in the new augmented feature space. The new augmented features represent general-features, source-specific features, and target-specific features. General-features are assumed to be shared between domains.
            In the new source augmented feature set, the target-specific features are zero and similarly, in the new target feature set, the source-specific features are zero, i.e.,
            \begin{gather*}
                \varphi_s(x)=<x, x, 0> \;\; \textrm{and} \;\; \varphi_t(x)=<0,x,x>,
            \end{gather*}
            where $\varphi_s(x)$ and $\varphi_t(x)$  are the mapping functions to the new feature spaces.
            Mapping the original features into a high-dimensional feature space separates the source features from the target features and gives the classifier the ability to learn the feature weights optimally. This method also can be easily applied to multiple source transfer learning.
            \cite{long2013transfer} proposed Joint domain adaptation (JDA) to match the domains where there is a discrepancy in both marginal and conditional distributions. JDA first reduces the dimensionality of original features using Principle Component Analysis (PCA) to discover the robust representation of data. It then embeds the extracted features into a higher dimension to match the marginal distributions using Maximum Mean Discrepancy (MMD). However, aligning the conditional distributions requires the labeled data in both source and target domains. In the absence of labeled data in the target domain, pseudo labels are usually used. A classifier trained on the source labeled data can be applied to predict the pseudo labels for the unlabeled data in the target domain. The model then utilizes the pseudo labels to minimize the conditional distribution divergence by modifying MMD. 
            JDA performs this process iteratively until it finds an optimal mapping function where both marginal and conditional distributions of the projected data can be jointly aligned. Ultimately, the target classifier can be trained on the domain-independent features extracted by the modified algorithm.
    \subsubsection{Parameter-Based}
        Parameter-based approaches assume that the source and target tasks comprise related knowledge. In this category of approaches, the model learns its parameters from the labeled source data and shares them with the target model. 
        Single-Model Knowledge Transfer (SMKT) \cite{tommasi2009more} is an SVM-based model that was proposed to address the object category problem. SMKT aims to learn a decision function from very limited labeled instances in the target domain and the transferred knowledge contained in the parameters of a pre-trained model to predict a new category in the target domain.
        If the source and target categories are related, the model performance will be improved significantly when the number of known categories in the source domain increases. However, if the learned categories differ from the new category, the model does not negatively impact the performance since the model uses a weighting strategy to transfer only the related knowledge to the target decision function.
        The weighting mechanism assigns proper weights to each source domain data based on their relatedness to the target domain through a leave-out-one cross-validation process \cite{cawley2006leave}. To improve the target classifier's performance, SMKT only transfers the knowledge of a single source domain in which its model parameters receive the highest weights.
        Multi-model knowledge transfer (MMKT) \cite{tommasi2010safety} extends SMKT to benefit from all pre-trained decision functions and exploit all useful prior knowledge.
        The idea behind this technique is that transferring knowledge from multiple source domains can boost the performance in the target classifier. For instance, if we want to learn a task from a set of know categories including cat, dog, bicycle, and car to classify motorbike, the model is expected to perform better when transferring the knowledge contained in both the car and bicycle than utilizing only one of them. 
        MMKT transfers prior knowledge by choosing a subset of known categories and controls the amount of transferred information from each category by finding the proper weighting parameter for the model.
    \subsubsection{Relational-Based}
        Relational-based transfer learning tends to discover the relationship between data in the source domain and transfer this relational knowledge to the target task. This approach is helpful when the sample data are not independently and identically distributed.
        \cite{li2012cross} proposes a domain adaptation method for the text classification problem where there is no labeled data in the target domain, while a large amount of labeled data is available in the source domain. Note that the source domain labels indicate each word as a sentiment word, topic word, neither sentiment nor topic word. This technique learns the relational knowledge contained in the source labeled data during the training stage and leverage them to help the learner precisely classify the words in the target domain into three classes of sentiments, topics, or neither.
        The proposed framework consists of two steps. In the first step, the model discovers the lexicons from the annotated data in the source domain. Lexicons include sentiment seeds, topic seeds, and structural patterns between them. This process extracts the common sentiment words between source and target domains. These words are termed as sentiment seeds and aim to bridge domains by utilizing a scoring metric. The scores are computed for each sentiment seed based on the probability of its occurrence in both domains.
        Next, the model extracts all the patterns between the sentiment and topic words in the source domain as candidate patterns to find the general patterns. Scores are then computed for the candidate patterns such that the patterns that are precise in the source domain and frequently occur in the target domain receive higher scores. The top-scored structural patterns and the extracted sentiment seeds are used to score the topic words in the source domain, and top-scored topic words are considered topic seeds.
        In the second step, the algorithm utilizes a Boostrapping-based method defined as a relational adaptive Boostrapping algorithm (RAP) to utilize the knowledge extracted in the first step in the target domain and help the classifier predict the labels for the target unlabeled data accurately. 
        RAP uses Transfer AdaBoost (TrAdaBoos) \cite{dai2007boosting} as a cross-domain classifier, aiming to reduce the negative transfer by re-weighting the source instances iteratively. In each iteration, two classifiers are separately trained on the existing lexicons, one for the sentiment seeds and one for the topic seeds, to predict the target labels. The algorithm then selects the top-scored sentiment and topic words as candidates and uses them along with the target seeds predicted in the previous iterations to construct a bipartite graph to score the candidate labels. The top-scored sentiment and topic words are added to the lexicons for the next iteration. Note that in the first iteration, the lexicons only consist of the sentiment and topic seeds found from the source domain in the first step. This process continues over an iteration number defined by the user.
\subsection{Heterogeneous Transfer Learning}
    Heterogeneous transfer learning is used to deal with the situation where the feature or/and label spaces vary between domains,i.e., $X_s \neq X_t$ and/or $Y_s \neq Y_t$. In these problem settings, the source and target domains share no features/labels directly. However, related knowledge in the source domain is supposed to be represented differently in the target domain.
    Heterogeneous transfer learning approaches mainly employ feature-based techniques to transfer knowledge across domains. In this category of approaches, transforming the source features/labels into the target domain is more challenging since the domains' features/labels originate from different spaces.
    Similar to homogeneous feature-based approaches, feature transformation can be accomplished symmetrically or asymmetrically. Symmetric feature transformation maps the source and target features into a common latent feature space, while asymmetric feature transformation maps the features from one feature space into another. Feature mapping techniques usually utilize pre-processing, dimensionality reduction, or/and feature selection methods.
    \subsubsection{Symmetric Feature-Based}
        As explained earlier, Feature Augmentation Method (FAM) \cite{daume2009frustratingly} works best for homogeneous transfer learning, where the source and target domains have the same feature space. However, replicating the different feature representations would be less effective in heterogeneous transfer learning. 
        \cite{duan2012learning} proposed the extended version of FAM called Heterogeneous Feature Augmentation (HFA). In this method, the source and target features are transformed into a common subspace by utilizing two projection matrices P and Q. New instances are then created by utilizing feature replication similar to FAM. The new source and target instances can be defined as follows,
        \begin{gather*}
            \varphi_s(x^s)=<Px^s, x^s, 0_{d_s}> \; \textrm{and} \;\; \varphi_t(x^t)=<Qx^t, 0_{d_t}, x^t>,
        \end{gather*}
        where $Px^s$ and $Qx^t$ are transformed source and target features in the common feature space and have the same dimension. $x^s$ and $x^t$ are the original source and target features. $0_{d_s}$ and $0_{d_t}$ are zero vectors with the same dimension as the original source and target features.
        Cross-Domain Landmark Selection (CDLS) \cite{hubert2016learning} was proposed to solve the semi-supervised heterogeneous transfer learning problems. It aims to learn invariant features across domains and identify the representative cross-domain landmarks. CDLS first projects the target samples into an m-dimensional subspace using PCA where $m \leq min{d_s, d_t}$. The reason for choosing m as the minimum value of the source and target dimensions is to avoid overfitting from mapping low dimensional data into a high dimensional space. The low-dimensional space is supposed to extract the underlying structure of the target data. It then learns a linear transformation $A$ to project the source data into the derived subspace. The model aims to match the marginal and conditional distributions in the common subspace using MMD. Parameters $\alpha$ and $\beta$ are considered landmark weights for the labeled source and unlabeled target instances, respectively.  The model computes $\alpha$ and $\beta$ during the conditional distribution alignment. The more similar an instance is to the target labeled data, the higher weight it will receive.
        A linear SVM learns from the labeled source and target data to predict pseudo labels for the unlabeled target data. All the instances with non-zero weight can be utilized to learn SVM. CDLS iteratively performs the above steps to update transformation $A$ and parameters  $\alpha$ and $\beta$ until it converges. 
    \subsubsection{Asymmetric Feature-Based}
        \cite{hoffman2013efficient} proposed an adaptation method for asymmetric feature-based heterogeneous transfer learning in multi-class image classification, where there exists a limited amount of annotated data in the target domain along with a large number of source labeled data. This method is termed as Max-Margin Domain transforms (MMDT). 
        MMDT simultaneously learns a linear transformation to map the target features into the source domain linearly. It adopts a multi-class SVM-based classifier to maximally separate the classes in the training set consisting of the source domain samples and the projected target samples. This adaptation method jointly optimizes both classifier parameters and the projection matrix to gradually learn a new target feature representation shared across multiple classes.
        The proposed method utilizes the linear transformation instead of using a similarity constraint. Therefore, the algorithm can be solved in linear feature space, making it computationally fast and scalable. This technique can be applied efficiently for the classification tasks where there are many training samples.
        Feature-Space Remapping (FSR) \cite{feuz2015transfer} aims to deal with the situations where the source and target domains are originated from different feature spaces. 
        FSR maps the target features to the source feature space using an asymmetric transformation strategy. It finds the mapping with the help of meta-features. The available target labels are utilized to create meta-features from the source and target instances. The model uses meta-features to construct a similarity matrix, containing the similarity scores computed for each pair of source and target features in the source domain space. 
        FSR then uses the feature pairs with the highest similarity scores to construct the mapping function. This mapping is supposed to minimize the error of the pre-trained source model when applied to the mapped target data.  
        This model forms a many-to-one mapping since various target features can be mapped into a single source feature. Therefore, FSR can simply use an aggregation procedure such as minimum, maximum, average, or total to combine the multiple dimensions produced by many-to-one mapping. 
\section{Application}
    Transfer learning is an exciting machine learning topic that has been recently applied in a wide variety of tasks and applications. In this section, we briefly discuss some of the real-world applications that have benefited from transfer learning.
    \cite{chattopadhyay2012multisource} proposed a transfer learning and domain adaptation technique for medical data. this technique is used for classification tasks to detect different stages of muscle fatigue based on the surface electromyogram data collected from different sensors.
    \cite{pan2010domain} introduced a method for indoor WiFi localization and cross-domain text classification tasks. The work proposed by \cite{li2012cross} also is used for text classification.
    \cite{daume2009frustratingly} proposed a technique for NLP applications. The technique can be applied specifically in labeling tasks such as named-entity recognition, shallow parsing, and part-of-speech tagging.
    Transfer learning can be also employed for image classification and object detection \cite{tommasi2009more}, \cite{tommasi2010safety}, \cite{hoffman2013efficient}, \cite{duan2012learning}.
    There are many other applications for transfer learning including reinforcement learning \cite{taylor2007cross}, metric learning \cite{zhang2010transfer}, text clustering \cite{gu2009learning}, image clustering \cite{yang2009heterogeneous}, sensor based location estimation \cite{zheng2008transferring}, collaborative filtering \cite{cao2010transfer}, etc.
%
\section{Conclusion}
    This paper clearly defines a machine learning methodology named transfer learning and its categorization from different perspectives. 
    We can view and classify transfer learning based on the availability of labeled data or the similarity/dissimilarity of source and target feature spaces.
    
    Besides, we briefly discuss some machine learning solutions to transfer learning problems. These approaches usually utilize instance-based, feature-based, parameter-based, and relational-based methods to extract knowledge in the source domain and leverage it into the target domain. 
    
    Ultimately, we concisely introduce some real-world transfer learning applications, including text classification, image classification, clustering, object detection, reinforcement learning, metric learning, etc.
    
\def\IEEEbibitemsep{0pt plus .5pt}

\end{document}